# A Method of Generating Measurable Panoramic Image for Indoor Mobile Measurement System

Hao Ma[1], Jingbin Liu[1, *], Zhirong Hu[2], Hongyu Qiu[1], Dong Xu[1], Zemin Wang[1], Xiaodong Gong[1], Sheng Yang[1]

[1]State Key Laboratory of Information Engineering in Surverying, Mapping and Remote Sensing, Wuhan University, Wuhan 430079, China - {mahao_fido, jingbin.liu, qiuhongyu, Whusggxudong, zeminwang, gongxiaodong, shengy}@whu.edu.cn

[2]Xi Rui Optoelectronics Technology Co., Ltd. - zrhu@xroe.net.cn

**Abstract:** This paper designs a technique route to generate high-quality panoramic image with depth information, which involves two critical research hotspots: fusion of LiDAR and image data and image stitching. For the fusion of 3D points and image data, since a sparse depth map can be firstly generated by projecting LiDAR point onto the RGB image plane based on our reliable calibrated and synchronized sensors, we adopt a parameter self-adaptive framework to produce 2D dense depth map. For image stitching, optimal seamline for the overlapping area is searched using a graph-cuts-based method to alleviate the geometric influence and image blending based on the pyramid multi-band is utilized to eliminate the photometric effects near the stitching line. Since each pixel is associated with a depth value, we design this depth value as a radius in the spherical projection which can further project the panoramic image to the world coordinate and consequently produces a high-quality measurable panoramic image. The purposed method is tested on the data from our data collection platform and presents a satisfactory application prospects.

**Keywords:** Image Stitching; Multi-Source Data Fusion; Graph Cuts

## 1. Introduction

With the development of smart cities, the demand for indoor 3D information is increasing. Compared with outdoors, indoors lack effective positioning methods, which makes it more difficult to obtain indoor 3D information. In order to solve this problem, the indoor mobile measurement system that integrates various sensors such as an inertial measurement unit (IMU), fisheye cameras, and LiDAR emerges as times require[1, 2]. This system can provide a variety of data sources, including fisheye images, 3D laser scanning point clouds, and further measure the scene by producing panoromic image and colored point clouds after data processing. Potential applications of the system lie in analyzing Urban and medical image, monitoring of large public places such as airports and train stations.

As an important product of the indoor mobile measurement system, Panoramic image provides 360 degree scanning of our surroundings. Being a key technology for generating panoramic image, Image stitching is an critical branch in the filed of digital image processing and computer vision, and is also know as image mosaic [3, 4]. The goal of image stitching is to create a wide-view, high-resolution and visually satisfactory image free of artifacts that may occur due to relative camera motion, illumination changes and optical aberrations, and the main process of image stitching consists of feature matching, registration/image alignment and seam removal [5, 6]. The purpose of feature matching is to find the correspond points of the images for stitching, image registration aligns the input images to the same coordinate system, and seam removal is employed to eliminate the seam to finally obtain a natural-looking mosaics [7, 8]. In this paper, image registration has been performed under given geometric transformation, so we mainly focus on optimal seam finding.

Geometric misalignment between input source images always exits because hardware related calibration is a tough task. With the presence of geometric misalignment, the obvious parallax can be solved by detecting the optimal seamline avoiding crossing majority of visually obvious objects and most of overlap regions with low image similarity. The optimal seamline is found by optimizing an objective function that minimizes the difference in the vicinity of the overlapping area [9, 10]. This involves the design of energy functions and the optimization of the solution. The energy functions are naturally defined by utilizing color, gradient and even texture, and are optimized through typical algorithms, such as snake model [11], dynamic programming [12] and graph cuts [13]. If the seamline and stitching artifacts are still visible due to color differences, the image blending technique can be further applied to solve it easily[14, 15]. Li [16] proposed a novel algorithm to efficiently detect optimal seamline for mosaicking aerial images captured from different viewpoints and for mosaicking street-view panoramic images without a precisely common center in a graph cuts energy minimization framework. This method fully utilized the information of image color, gradient magnitude and texture complexity into data and smooth every terms in graph cuts. In terms of our problem during our application, seamline detection plays a vital role. We choose this novel method due to its high-quality seamline for multiple image mosaicking. We have realized the algorithm using C++ and achieved an satisfactory results.

To generate measurable panoramic image, LiDAR and image data are integrated together to obtain dense depth map according to the transformation between these two sensors. Basically, dense depth map which is in pixel-wise compactness with RGB image is upsampled from a sparse depth map, which is obtained by downsampling a dense depth map and either the depth value of each pixel is calculated within a local patch or a global energy function is constructed to estimate depth value of all pixels simultaneously [17, 18]. Under the assumption that the LiDAR and image are well aligned, the key problem is to design a upsampling method to keep the boundary of the resultant dense depth map consistent with that of RGB image and get rid of depth inhomogeneity [19, 20]. however, upsampling a sparse map obtained through a simple dynamic projection from 3D LiDAR point to 2D image plane is particularly challenging. Chen [18] described these difficulties that need to be solved during the depth upsampling and designed a self-adaptive method to upsample the sparse depth map in which the RGB image and the anisotropic diffusion tensor are exploited to guide upsampling by reinforcing the RGB-depth compactness, and global enhancement is adopted in the final step [21]. According to the excellent performance in both indoor and outdoor scenarios, we select this method to generate our dense depth map.

The remaining part of this paper is organized as follows. Fusion of LiDAR and image data is firstly presented to make the produce process of dense depth map clear. Then, we introduce the graph-cuts-based image stitching method. Next, we combine the image stitching and depth map to obtain a measurable panoramic image. Experiments and evaluations on indoor data are elaborately designed and conclusions are finally drawn.

## 2. Fusion of LiDAR and image data

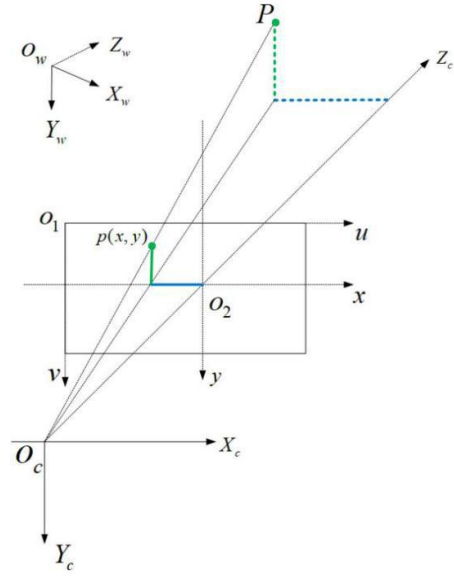

Fig. 1. Camera coordinate and world coordinate

Transformation between 3D LiDAR scanner and a CCD camera involves several coordinates. Suppose the coordinate of 3D point cloud is $O_w$-$X_wY_wZ_w$ and $O_c$-$X_cY_cZ_c$ is the camera coordinate. $O_1$-uv is the pixel coordinate and $O_2$-xy is the image plane coordinate. p(x, y) on the image plane is the projection of P. Above mentioned coordiantes can be illustrated together in Fig. 1:

and their mathematical relationship is:

$$P_C = RP_W + t \qquad (1)$$

where $R$ and $t$ are the rotation and translation between word and camera coordinate. Subscript in P denotes specific coordinate: 'C' means the camera coordinate and 'W' the world coordinate. The formulation above is equivalent to:

$$\begin{bmatrix} X_C \\ Y_C \\ Z_C \\ 1 \end{bmatrix} = \begin{bmatrix} R & t \\ 0 & 1 \end{bmatrix} \begin{bmatrix} X_W \\ Y_W \\ Z_W \\ 1 \end{bmatrix} \qquad (2)$$

further, since:

$$\begin{cases} x = f \cdot X_C / Z_C \\ y = f \cdot Y_C / Z_C \\ u = x/dx + u_0 \\ v = y/dy + v_0 \end{cases} \qquad (3)$$

let

$$\begin{cases} f_x = f/dx \\ f_y = f/dy \end{cases} \qquad (4)$$

where ($u_0$, $v_0$) are the coordinate of the principal point. $dx$ means and $dy$ means physical sizes of one pixel in horizontal and vertical direction respectively. The corresponding relationship between pixel

and word coordinate can be finally obtained:

$$Z_C \begin{bmatrix} u \\ v \\ 1 \end{bmatrix} = \begin{bmatrix} f_x & 0 & u_0 & 0 \\ 0 & f_y & v_0 & 0 \\ 0 & 0 & 1 & 0 \end{bmatrix} \begin{bmatrix} R & T \\ 0 & 1 \end{bmatrix} \begin{bmatrix} X_w \\ Y_w \\ Z_w \\ 1 \end{bmatrix} \quad (5)$$

Now, we have 3D point cloud, image data and transformation between them, we can generate depth map next. Firstly, point cloud can be projected on to image plane to generate initially sparse depth map. To make it more clear and direct, we have project point clouds onto the image plane in Fig. 2 and the valid locations are green-colored under the condition that the LiDAR and camera are preliminary calibrated and synchronized beforehand. If more than one 3D point clouds correspond to one pixel, the smallest depth is selected to avoid the occlusion existing in the scene structure and assigned to the sparse depth map in the same location with the pixel. Our original images captured by fisheye camera have been corrected using OcamCalib proposed by Scaramuzza [22], which is a famous correction tool embedded in Matlab software. We don't describe the correction method here due to space limitations, and interested readers are recommended to read Scaramuzza's paper.

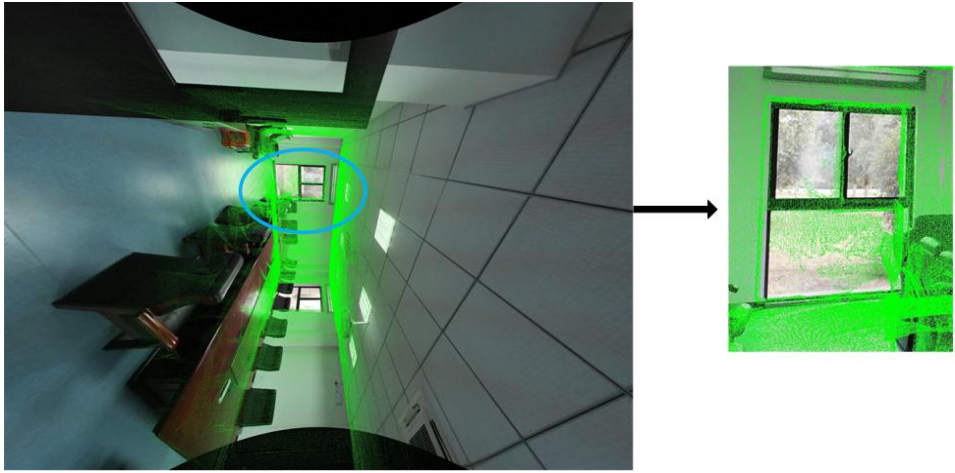

Fig. 2. Projection result from 3D point cloud to 2D image plane

Abstracted by the excellent performance in keeping consistent with RGB image, we adopt a self-adaptive method to generate a dense depth map [18]. Fig. 3 taken from Chen's paper presents performance of different methods. As we can see from Fig. 3(e), the boundary of dense depth map is well consistent with that of RGB image, which is a vital factor to judge the quality of the depth map generation algorithm. More details about this method can be found in Chen's paper.

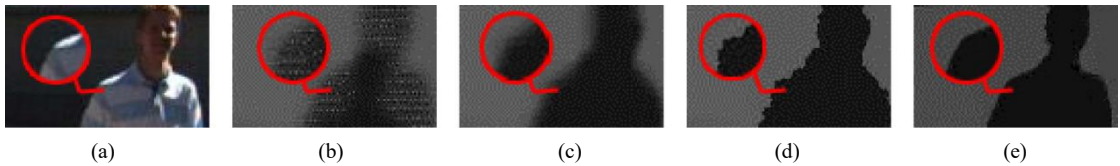

(a)     (b)     (c)     (d)     (e)

Fig. 3. Visual comparison of depth upsampling result under various conditions. (a) RGB image; (b) dense depth map without depth uncertainty elimination; (c) dense depth map with empirical kernel bandwidth; (d) dense depth map with self-adaptive bandwidth; (e) dense depth map with the global enhancement after self-adaptive bandwidth

## 3. Image stitching

*3.1 Data collection and definition of coordiantes*

Our data collection platform is armed with LiDAR, fisheye camera and IMU. LiDAR and IMU can output location and posture under the condition without GNSS and fisheye cameras take pictures of surroundings when the platform is on the move. Sensors are elaborately calibrated into a uniform coordinate system. 3D point clouds is generated using simultaneous localization and mapping (SLAM) and the pose of captured images can be calculated using SLAM and calibration parameters obtained in advance. Now, it is enough to start generating depth map and panoramic image under the assumption that images have been geometrically aligned as precisely as possible.

The layout of our cameras is shown in Fig. 4. Panoramic image processing involves image plane coordinate system, camera coordinate system ($S_i$, where $i$ is the camera index), virtual image space coordinate system Sv-XYZ, point cloud coordinate system (world coordinate system), all of which are right-handed coordinate systems. The origin of the virtual image space is located at the geometric center of the six fisheye cameras, and the orientation is consistent with that of the mobile measuring platform equipped with the fisheye cameras. All coordinate systems are registered to the point cloud coordinate system. Based on the depth value, the 2D point in the image coordinate system is transferred to the 3D coordinate in the point cloud coordinate system, and then the coordinates are converted to the virtual image space coordinate system which is further registered to the panoramic image through spherical projection.

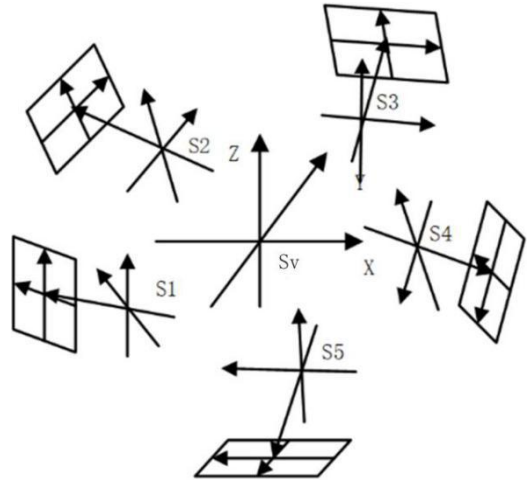

Fig. 4. Camera coordinates and virtual image space coordinate

*3.2 Projection model*

There are two kinds of mapping models between the spherical surface and the panoramic image: direct mapping and inverse mapping. In general case, the radius of the spherical surface is a fixed value, while each pixel is associated with a depth value in our paper, and the radius is set as the pixel's depth. Specifically, for direct mapping, each image pixel is projected onto the spherical surface and the spherical surface is mapped to the panoramic image. Inverse mapping traverses the panoramic image pixels using the generated panoramic image depth map as the spherical radius, generates the corresponding three-dimensional point coordinates, and finally project it onto the fisheye image. The direct mapping will result in empty points in the panoramic image, so the inverse mapping is adopted to generate the panoramic image, and the depth maps corresponding to the six cameras are utilized to generate the panoramic depth map.

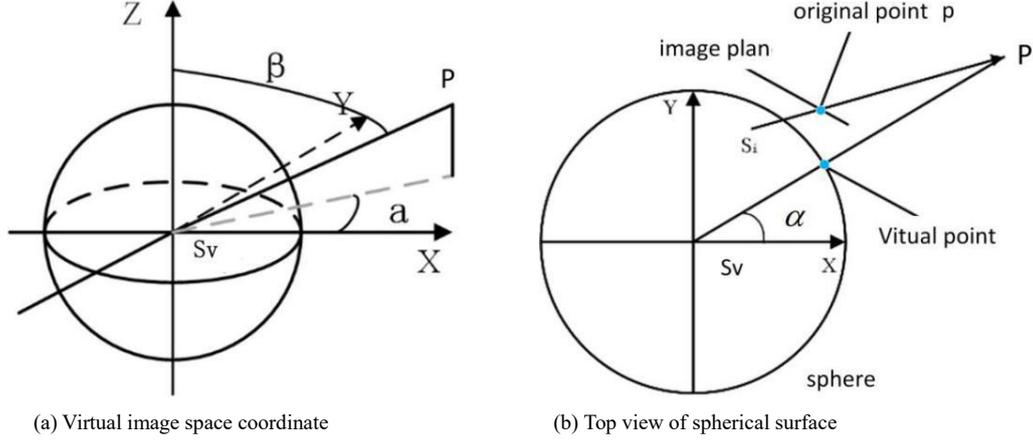

(a) Virtual image space coordinate  (b) Top view of spherical surface

Fig. 5. Virtual image space coordinate

In Fig. 5, Sv is the origin of the virtual image space coordinate, and ($X_v$, $Y_v$, $Z_v$) represents point P on an object in the scene structure in terms of the virtual coordinate. α is a angle between line Sv-P and the plane Sv-XY, β is a angle between line Sv-P and the Z axis, and they are in [-π, π] and [0, π] respectively. If the height and width of panoramic image are V and U, the corresponding point of ($X_v$, $Y_v$, $Z_v$) on the panoramic image is:

$$\begin{cases} u_{pano} = U*(\pi - \arctan\dfrac{Y_v}{X_v})/2\pi \\ v_{pano} = V - (\arctan\dfrac{\sqrt{X_v^2+Y_v^2}}{Z_v} + \dfrac{\pi}{2})/V \end{cases} \quad (6)$$

The inverse map searches pixel on the undistorted or fisheye image for every pixel on the panoramic image plane. For any pixel $p'(u_{pano}, v_{pano})$, since

$$\begin{cases} \alpha = \pi - (u_{pano} \cdot 2\pi/U) \\ \beta = \pi * v_{pano}/V \end{cases} \quad (7)$$

once we know the depth $D$ of $P_v$, then,

$$\begin{cases} X_v = D\cdot\sin\beta\cos\alpha \\ X_v = D\cdot\sin\beta\sin\alpha \\ Z_v = D\cdot\cos\beta \end{cases} \quad (8)$$

Further, there exists an relationship between undistorted image plane and the virtual coordinate. Set $R_v$ and $T_v$ are rotation and translation between the virtual image space coordinate and the world coordinate:

$$P_v = R_v P_w + T_v \quad (9)$$

it is equivalent to

$$P_w = R_v^{-1}(P_v - T_v) \quad (10)$$

according equation (5), we have

$$Z_s^p \cdot p = K_s(R_s P_w + T_s) \quad (11)$$

where $R_s$ and $T_s$ are rotation and translation between the word coordinate and camera coordinate system. $Z_s^p$ is the depth of point P on a object and $K_s$ is the intrinsic parameters in term of a specific camera. p is the projection of P on the image plane. So once we have the depth of a pixel on the panoramic image, we can obtain its corresponding pixel on the undistorted image and vice versa. Given an image from one camera, it appears in an area of the panoramic plane through the inverse projection, as shown in Fig. 6.

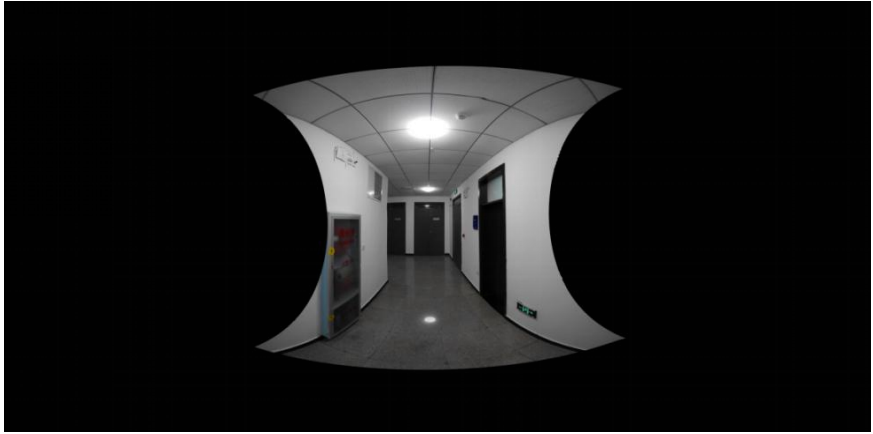

Fig. 6. Inverse projection

To generate the panoramic depth map, direct mapping described above can directly project undistorted image plane to spherical surface, then to panoramic plane. We can calculate depth map for each undistorted image using the method mentioned in Section 2 beforehand. After direct mapping, average value is adopted for overlap area in the panoramic plane. For pixels with invalid depth values in the panoramic plane, we fill it using the smallest depth value from its surrounding valid values. Specifically, eight directions are considered here: Up, down, left, right, northeast, northwest, southeast, southwest and the first valid one is searched in each direction, then the smallest one among these valid values is given to the invalid. Here, the bottom of the panoramic depth map is ignored and will be described in section 3.3.2.

With the panoramic image and depth map, we can measure any physical size of the object in the real world. Specifically, if $p_1$ and $p_2$ are two different points on the generated panoramic image, we can deduce the their real distance in term of the world coordinate utilizing the panoramic depth map. Let $D_1$ and $D_2$ are the depth values of $p_1$ and $p_2$ respectively, then their three-dimensional coordinates can be calculated through equations (7)-(10) and the euclidean distance of the two 3D coordinates is the desired distance.

### *3.3 panoramic image*
### *3.3.1 Graph-cuts-based seamline searching*

Because of geometric misalignment, after projecting images of one station to the panoramic plane, seamline needs to be searched in the overlap area to make a visually natural and pleasant result. We adopt Li's graph-cuts-based seamline searching method because of its excellent performance [16]. To build an graph for image stitching, each pixel is viewed as a vertex in the graph, and weight of a edge between vertexes is inversely proportional to the possibility the two vertex should be cut. There are two virtual terminals in the graph, and each pixel in the image should be connected to the terminals. The edge between ordinary vertexes is called n-links and the edge between ordinary vertex and the terminals t-links.

Two-label graph cuts optimization example is shown in Fig. 7 where A and B are terminals. Fig. 7 is taken from Li [16]. Seamline should go through pixels that show a slight difference in terms of color and gradient. Set color $C_c(\mathbf{x})$ and gradient $C_g(\mathbf{x})$ difference as the energy of a vertex and the sum of the two vertexes' energy as the weight on this edge. Specifically, $\mathbf{x}$ represents a pixel on the overlap area. RGB images is conversed to HSV space ahead, $I_1$ and $I_2$ are two images and $w$ is a weight to balance the difference in $V$ channel and $S$ channel.

$$C_C(\mathbf{x}) = w|V_{I_1}(\mathbf{x}) - V_{I_2}(\mathbf{x})| + (1-w)|S_{I_1}(\mathbf{x}) - S_{I_2}(\mathbf{x})| \tag{12}$$

subscript in $V$ and $S$ means a specific image. For gradient-based term, the optimization direction should prevent seamline going through pixels with big gradient and big gradient difference, because big gradient happens in areas with big visually image split and the small one in areas with natural transition.

$$C_g(\mathbf{x}) = \frac{1}{4}(|G^x_{I_1}(\mathbf{x})| + |G^x_{I_2}(\mathbf{x})| + |G^y_{I_1}(\mathbf{x})| + |G^y_{I_2}(\mathbf{x})|) \\ + |G^x_{I_1}(\mathbf{x}) - G^x_{I_2}(\mathbf{x})| + |G^y_{I_1}(\mathbf{x}) - G^y_{I_2}(\mathbf{x})| \tag{13}$$

Superscript in $G$ means direction of gradient and subscript means a specific image. Finally, combining these two term together, we can get energy for each pixel in the overlap area

$$C(\mathbf{x}) = C_C(\mathbf{x}) + C_g(\mathbf{x}) \tag{14}$$

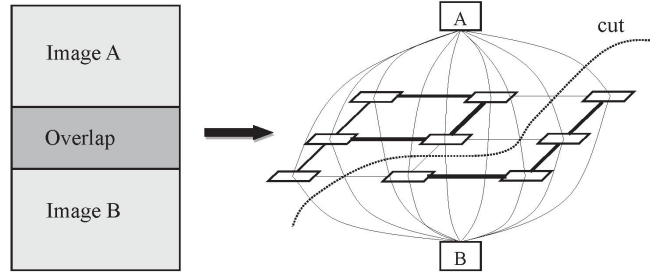

Fig. 7. Illustration of graph-cuts-based seamline searching

Compared with Li's graph-cuts-based seamline searching method, we ignore texture complexity term for simplicity, and this is feasible from the validation of our experiment. Next, we can design

$$E(I) = E_{data}(I) + E_{smooth}(I) \tag{15}$$

where $I$ denotes the overlap area in Fig. 7. Further

$$E_{data}(\mathbf{x}) = \begin{cases} 0 & \mathbf{x} \in I \\ \infty & \mathbf{x} \in I \end{cases} \tag{16}$$

and

$$E_{smooth}(\mathbf{x}, \mathbf{y}) = C(\mathbf{x}) + C(\mathbf{y}) \tag{17}$$

$\mathbf{y}$ is a neighbor vertex of $\mathbf{x}$ in the graph. In our application, overlap area $I$ is the intersection of two minimum bounding rectangle of two images to be stitched (as shown in Fig. 6, image is projected to a panoramic plane while the appearance of valid projected area is irregular. The valid projected area is considered to calculate the minimum rectangle). Under this situation, for one pixel **x** in the overlap area *I*, it has four cases: 1) **x** belongs to $I_1$ but doesn't belongs to $I_2$; 2) **x** belongs to $I_2$ but not belongs to $I_1$; 3) **x** belongs to $I_1$ and $I_2$; 4) **x** doesn't belong to $I_1$ and $I_2$. So, to make it more clear, we define

$$E_{smooth}(\mathbf{x},\mathbf{y}) = \begin{cases} C(\mathbf{x})+C(\mathbf{y}) & \mathbf{x} \in I_1 \cap I_2,\ \mathbf{y} \in I_1 \cap I_2 \\ P1 & \mathbf{x} \in I_1 - I_2\ or\ \mathbf{x} \in I_2 - I_1,\ \mathbf{y} \in I_1 \cap I_2 \\ P1 & \mathbf{y} \in I_1 - I_2\ or\ \mathbf{y} \in I_2 - I_1,\ \mathbf{x} \in I_1 \cap I_2 \\ P1 & \mathbf{x} \notin I_1 \cup I_2,\ \mathbf{y} \notin I_1 \cup I_2 \end{cases} \quad (18)$$

P1 can be set to a value that is two times larger than the biggest gradient in the image. With the presence of P1, the energy minimization makes seamline appear in the intersection of the valid projected area of the two images.

*3.3.2 Black hole filling*

Because of the absence of the bottom camera, the initial panoramic image has an black area. To fill this black area, images from neighboring stations should be considered. For a certain station, images captured by different cameras and its corresponding panoramic image have an relationship with current virtual coordinate, and different stations have their own transformation to the world coordinate, so pixel in each image and panoramic image can make a connection with that in other image or panoramic image. Since the black area in current station can always be seen in other station, it is possible to fill it using this strategy.

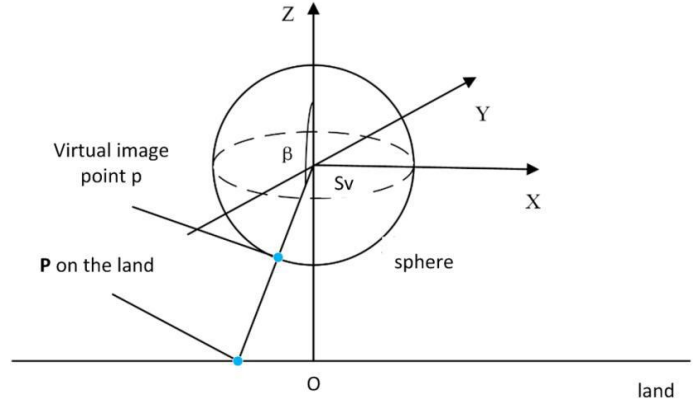

Fig. 8. Schematic diagram of the relationship between

the bottom hole and the virtual image space coordinate

Each black hole in the panoramic image basically has same size, and we notice that the vertical length of the black hole can be uniformly set to two-tenth of the height of the panoramic image. For pixel in the black hole, α and β can be calculated. If the length of SvO in Fig. 8 is *h*, the length of SvP is *D*, then

$$D = h / \cos \beta \quad (19)$$

once we know *D*, the coordinate of the pixel in the black hole can be further obtained in terms of the world coordinate according to equations (7)-(10), then corresponding pixels in other images which belong to other stations will be found naturally based on equations (9) and (6). Here, we search images in neighboring stations, and simply pick the first valid RGB to that in black hole due to our reliable calibrated parameters.

*3.3.3 Image blending*

Photometric inconsistencies appear on the two sides of the seamline. Image blending can compensate the influence and we adopt a multi-band blending strategy here [15]. A multi-band blending scheme ensures smooth transitions between images despite illumination differences, whilst preserving high frequency details. The idea behind multi-band blending is to blend low frequencies over a large spatial range, and high frequencies over a short range. This can be performed over multiple frequency bands using a Laplacian Pyramid. Interested readers can refer to [8, 15] for more detailed information.

## 4. Experiments

Fig. 9 shows the scene of our experiments. Red points are the tracks of our data collection platform. The experimental site is rich, which contains long corridor, hall and office.

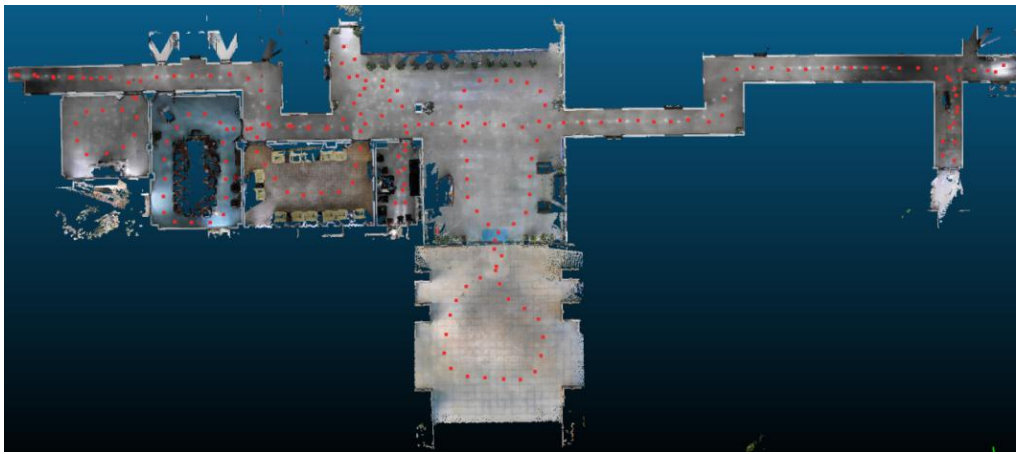

Fig. 9. Experimental site

The panoramic depth map of a certain station is firstly showed in Fig. 10(a). From the observation of Fig. 10(c), the panoramic depth map is basically consistent with the scene structure: the brighter the grayscale value is, the bigger the depth value is, and the further the object is from the camera. Fig. 10(b) shows a panoramic image without image blending and the found seamline are marked with red lines. Without the involvement of image blending, the color difference on both sides of the seamline is obvious. Fig. 10(c) is the final panoraic image after the multi-band blending. More experimental results are presented in Fig. 11.

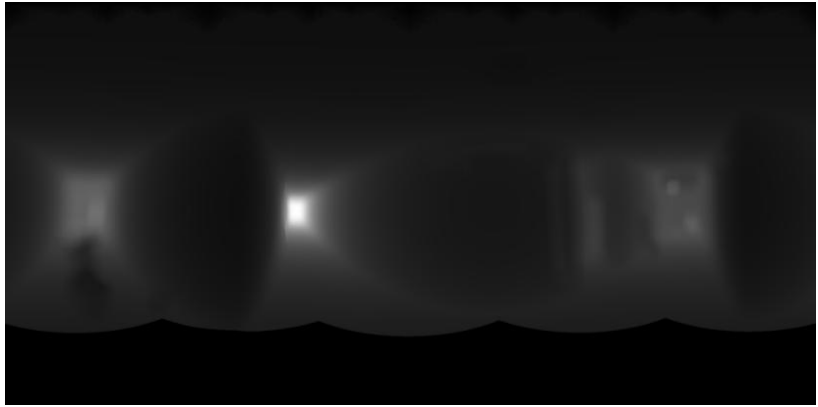

(a)

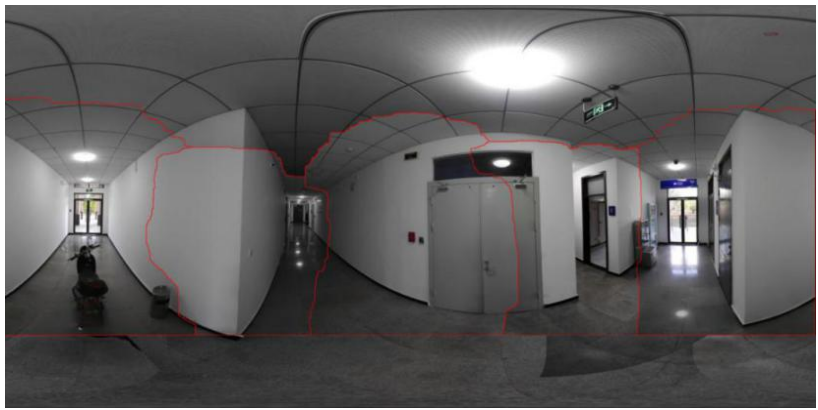

(b)

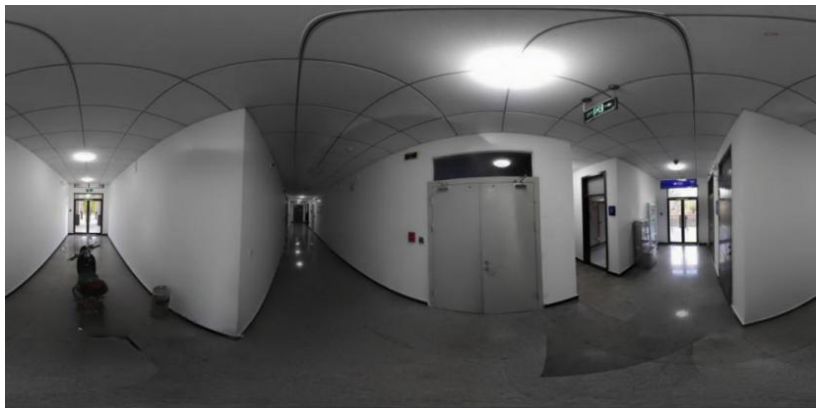

(c)

Fig. 10. Panoramic depth map and panoramic image. (a) Panoramic depth map.
(b) Panoramic image without image blending. (c) Final panoramic image.

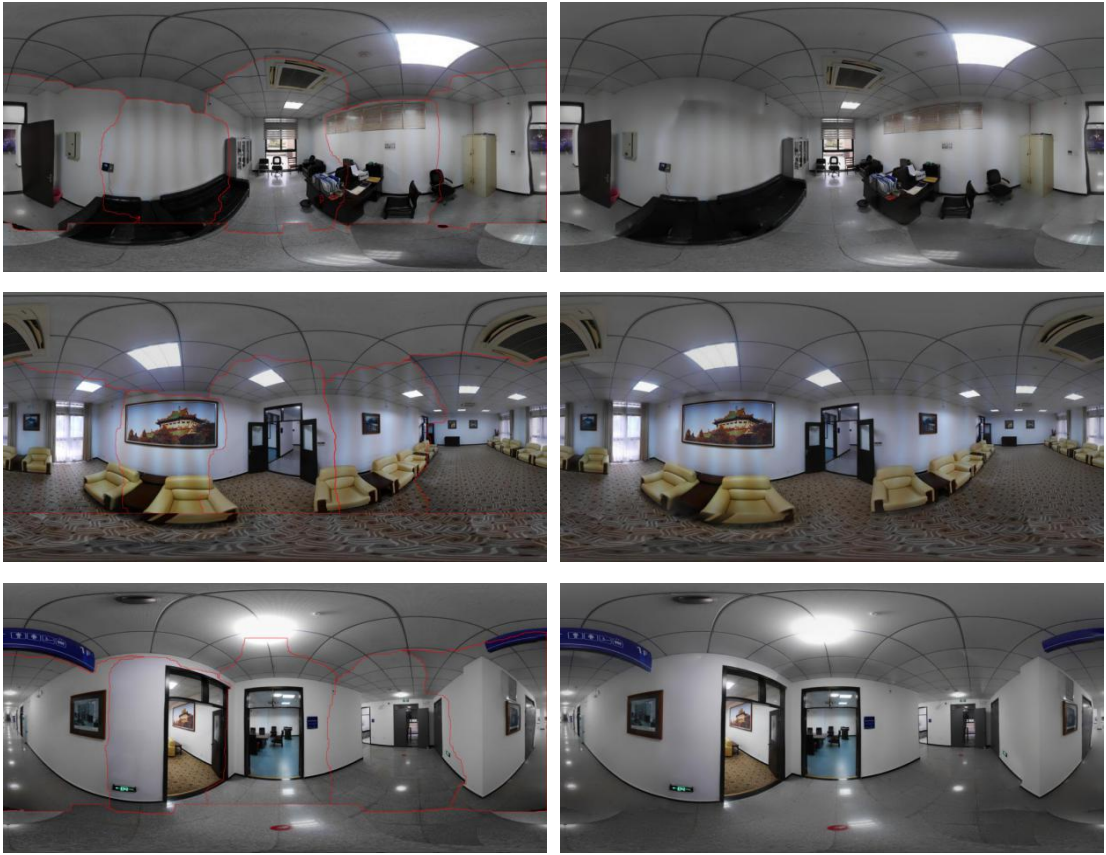

Fig. 11. Panoramic images

To evaluate the measurement accuracy, we randomly pick several lines from the panoramic image which are also straight lines in the real world as shown in Fig. 12. Although there is distortion in the panoramic image, what we care is the straight line distance. Number from ① to ⑨ are the index of the end point, and i in Li from 1 to 5 is the index of line. ($u_{pano}$, $v_{pano}$) is coordinate values in terms of pixel coordinate. 'D' represents the depth value, 'ML' the measured length of a line, 'GT' the ground truth of the line's length and 'Error' is the difference between 'ML' and 'GT'. The unit of these four quantities is meter. The measurement error for L1, L2, L3 and L5 is within a few centimeters and show a high accuracy. For line L4, a relative large error happens, this is because the end points' depth values should be close but there is an error of nearly 40cm. Naturally, the measurement accuracy is bonded closely with that of depth value, so the key is the fusion of LiDAR data and image data.

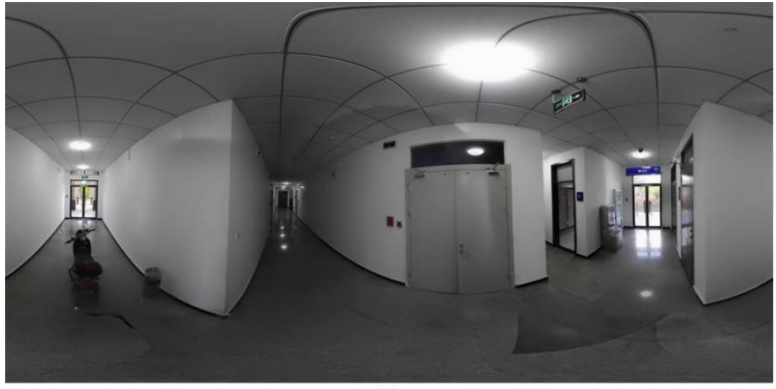

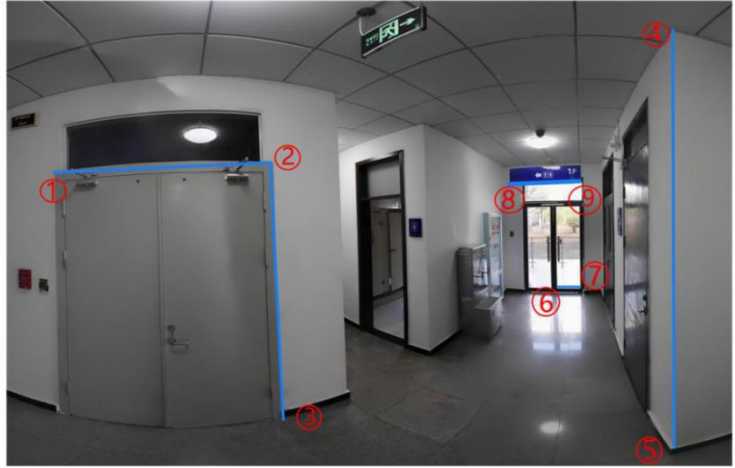

Fig. 12. Lines and points in the panoramic image

**Table. 1. Evaluation on the measurement accuracy**

|  | L1 | | L2 | | L3 | | L4 | | L5 | |
|---|---|---|---|---|---|---|---|---|---|---|
|  | ① | ② | ② | ③ | ⑥ | ⑦ | ⑧ | ⑨ | ④ | ⑤ |
| $u_{pano}$ | 4262 | 5394 | 5394 | 5442 | 6864 | 6986 | 6624 | 6986 | 7464 | 7469 |
| $v_{pano}$ | 1802 | 1748 | 1748 | 3076 | 2407 | 2409 | 1865 | 1850 | 1065 | 3236 |
| D(m) | 2.036 | 1.94 | 1.94 | 2.50 | 5.84 | 5.77 | 5.48 | 5.14 | 1.69 | 2.21 |
| $X_w$ | -3.214 | -1.592 | -1.592 | -1.672 | 3.001 | 2.870 | 2.868 | 2.482 | -1.452 | -1.446 |
| $Y_w$ | 2.006 | 1.763 | 1.763 | 1.695 | -0.235 | -0.746 | 0.696 | -0.746 | -0.529 | -0.496 |
| $Z_w$ | 2.325 | 2.324 | 2.324 | 0.106 | 0.073 | 0.069 | 2.465 | 2.440 | 2.955 | 0.054 |
| ML(m) | 1.64 | | 2.22 | | 0.53 | | 1.47 | | 2.90 | |
| GT(m) | 1.59 | | 2.21 | | 0.53 | | 1.25 | | 2.86 | |
| Error | 0.05 | | 0.01 | | 0 | | 0.22 | | 0.04 | |

## 5. Conclusions

In this article, we present a method to produce panoramic images integrated with depth information. Many work needs to be finished to achieve this goal: fusion of LiDAR and image data, image stitching, image blending and so on. An self-adaptive method is adopted in our paper to generate depth map while keeping the depth consistent with the RGB image. The resultant panoramic depth map consequently reflects the scene structure well. To search optimal seamline, graph-cuts-based method is utilized here and shows an satisfactory image stitching output. The final quality of the measurable panoramic image has also been validated using data collected from our platform. In the next work, we will design algorithm for fusion 3D point cloud and image data to improve the measurement accuracy, which depends largely on the depth value.

**Funding.** The Natural Science Fund of China with Project No. 41874031 and 61872431; the National Key Research Development Program of China with project No.2016YFB0502204 and 2016YFE0202300; the Technology Innovation Program of Hubei Province with Project No. 2018AAA070; the Natural Science Fund of Hubei Province with Project No. 2018CFA007.

**Disclosures.** The authors declare no conflicts of interest.